\documentclass{article}

\usepackage{microtype}
\usepackage{graphicx}
\usepackage{subfigure}
\usepackage{booktabs} 

\usepackage{hyperref}

\usepackage[accepted]{icml2024}

\usepackage{amsmath}
\usepackage{amssymb}
\usepackage{mathtools}
\usepackage{amsthm}
\usepackage{bm}
\usepackage{multirow}

\usepackage[capitalize,noabbrev]{cleveref}

\theoremstyle{plain}
\newtheorem{theorem}{Theorem}[section]

\theoremstyle{definition}
\newtheorem{definition}[theorem]{Definition}

\theoremstyle{remark}

\usepackage[textsize=tiny]{todonotes}

\icmltitlerunning{Submission and Formatting Instructions for ICML 2024}

\begin{document}

\twocolumn[
\icmltitle{Feature Attribution with Necessity and
Sufficiency\\via Dual-stage Perturbation Test for Causal Explanation}



\icmlsetsymbol{equal}{*}

\begin{icmlauthorlist}
\icmlauthor{Xuexin Chen}{gdut}
\icmlauthor{Ruichu Cai}{gdut,PZLab}
\icmlauthor{Zhengting Huang}{gdut}
\icmlauthor{Yuxuan Zhu}{gdut}
\icmlauthor{Julien Horwood}{cam}
\icmlauthor{Zhifeng Hao}{stu}
\icmlauthor{Zijian Li}{mbz}
\icmlauthor{José Miguel Hernández-Lobato}{cam}
\end{icmlauthorlist}

\icmlaffiliation{gdut}{School of Computer Science, Guangdong University of Technology, Guangzhou, China}
\icmlaffiliation{PZLab}{Pazhou Laboratory (Huangpu), Guangzhou, China}
\icmlaffiliation{cam}{Department of Engineering, University of Cambridge, Cambridge CB2 1PZ, United Kingdom}
\icmlaffiliation{mbz}{Mohamed bin Zayed University of Artificial Intelligence, Masdar City, Abu Dhabi}
\icmlaffiliation{stu}{Shantou University, Shantou 515063, China}

\icmlcorrespondingauthor{Ruichu Cai}{cairuichu@gmail.com}

\icmlkeywords{Machine Learning, ICML}

\vskip 0.3in
]



\printAffiliationsAndNotice{\icmlEqualContribution} 

\begin{abstract}
We investigate the problem of explainability for machine learning models, focusing on Feature Attribution Methods (FAMs) that evaluate feature importance through perturbation tests. Despite their utility, FAMs struggle to distinguish the contributions of different features, when their prediction changes are similar after perturbation. To enhance FAMs' discriminative power, we introduce Feature Attribution with Necessity and Sufficiency (FANS), which find a neighborhood of the input such that perturbing samples within this neighborhood have a high Probability of being Necessity and Sufficiency (PNS) cause for the change in predictions, and use this PNS as the importance of the feature. Specifically, FANS compute this PNS via a heuristic strategy for estimating the neighborhood and a perturbation test involving two stages (factual and interventional) for counterfactual reasoning. To generate counterfactual samples, we use a resampling-based approach on the observed samples to approximate the required conditional distribution. 
We demonstrate that FANS outperforms existing attribution methods on six benchmarks. 
Please refer to the source code via  \url{https://github.com/DMIRLAB-Group/FANS}.
\end{abstract}

\section{Introduction}
Feature attribution is a method for explaining machine learning (ML) models by assigning weights to input features, where the absolute value of these weights denotes their contribution to a model's prediction. Mathematically, suppose we have a function $f$ that denotes a trained ML model, a target input $\mathbf{x}^t$$=$$(x^t_1, \cdots, x^t_d)$ to be explained, and a baseline $\mathbf{x}'$$=$$(x'_1, \cdots, x'_d)$
, where $x'_i$ is the approximate value that feature $x^t_i$ would take if it were considered missing. An attribution of the prediction at input $\mathbf{x}^t$ relative to $\mathbf{x}'$ is a vector $\alpha_f(\mathbf{x}^t, \mathbf{x}')$$=$$(a_1, ..., a_d)$, where $a_i$ is the contribution of $x^t_i$ to the prediction $f(\mathbf{x}^t)$.

Standard feature attribution methods (FAMs) measure the contribution of each feature through a \emph{perturbation} test, i.e.,  
comparing the difference in prediction 
under different perturbations, e.g., Shapley Values~\cite{shapley1953value}, or before and after the perturbation, e.g., gradient or surrogate-based methods~\cite{lundberg2017unified,sundararajan2017axiomatic}, where a subset of the variables of $\mathbf{X}$ are set to their baseline values.

However, this perturbation test may not accurately distinguish the contribution of different features to the prediction when their changes in prediction are similar after perturbation. To better understand this phenomenon, we provide an example as follows.

\textbf{Example 1.}\label{exam:classifier}
Consider a binary classification model $f$ for  
$\mathbf{X}$$=(X_1,X_2,X_3)$ $\in$ $\mathbb R^3$ that is captured by a function $Y$=$h(X_1,X_2,X_3)$=$\mathbb I(X_1$$-$$X_2$$>$$1)$, where $Y$ is a binary variable representing true or false. Assume the baseline $\mathbf{x}'$$=$ $(0,0,0)$ and the target input $\mathbf{x}^t$$=$$(1,1,1)$.

The attribution scores for features $x^t_1$ and $x^t_3$ obtained from the existing FAMs may both be $0$, as the model's predictions given $\mathbf{x}^t$ under different perturbations relative to $\mathbf{x}'$ 
are always equal to $h(\mathbf{x}^t)$, which may misleadingly suggest that $x^t_1, x^t_3$ contribute equally to $h(\mathbf{x}^t)$.  
Clearly, the contributions of $x^t_1, x^t_3$ are not equal, since the value of $h(\mathbf{x}^t)$ is determined solely by the difference between $x^t_1$ and $x^t_2$ being less than $1$, and is not influenced by $x^t_3$.
In other words, the causes for the predictions for $x^t_1$ and $x^t_3$ remaining unchanged after perturbation are different. 
\begin{figure}[t]
	\centering
	\includegraphics[width=0.35\columnwidth]{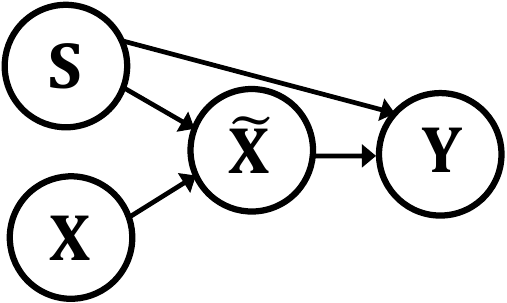} 
	\caption{Causal diagram of standard perturbation test in feature attribution. $\mathbf{S}$ denotes a subset of dimensions of $\mathbf{X}$ for perturbation. $\tilde{\mathbf{X}}$ represents an input with fixed features on $\mathbf{S}$ that are similar to the target input $\mathbf{x}^t$.}
        \label{fig:process}
\end{figure}
Based on this key observation, we extend feature attribution by finding a neighborhood of the input $\mathbf{x}^t$ containing $x^t_3$ (or $x^t_2$) such that perturbing samples within this neighborhood has the highest probability of causation called \emph{Probability of Necessity and Sufficiency} (PNS) of two counterfactual questions about $x^t_3$, and we take this PNS as the importance of $x^t_3$. The counterfactual questions are as follows. 
``\emph{for the inputs within this neighborhood, and the predictions for these inputs remain unchanged when some features (except $x^t_3$) are perturbed, would the predictions change if $x^t_3$ were perturbed?}'' and ``\emph{for the inputs within other neighborhoods containing $x^t_3$, and the predictions for these inputs change when only $x^t_3$ is perturbed, would the predictions remain unchanged if some features (except $x^t_3$) were perturbed?}'' The first question assesses the sufficiency of perturbing $x^t_3$ (cause A) for the prediction change (effect B), i.e., the ability of cause A to produce effect B. Similarly, the second question assesses the necessity, i.e., the dependence of effect B on cause A~\cite{2000Causality}. 
In Example \ref{exam:classifier}, the PNS for $x^t_3$ is 0, since perturbing $x^t_3$ is impossible to cause the prediction to change, 
while the highest PNS for $x^t_1$ may be greater than 0 because the prediction may change when $x^t_1$ is perturbed.

The above analysis demonstrates that evaluating the maximum PNS for prediction change more effectively distinguishes the importance of features than the extent of those changes. Hence, we develop a novel Feature Attribution with Necessity and Sufficiency (FANS) to compute the maximum PNS of the above two counterfactual questions about each feature subset. 
Specifically, we abstract the common understanding of the perturbation test in feature attribution by proposing a structural causal model (SCM). 
We then propose a heuristic strategy to calculate a neighborhood of $\mathbf{x}^t$ that is used to estimate the highest PNS of the counterfactual questions for each feature subset. 
To quantify this PNS, we then propose a novel dual-stage perturbation test to implement the counterfactual reasoning paradigm \emph{Abduction-Action-Prediction}~\cite{pearl2013structural}. The first stage is to generate samples from a distribution condition on the factual perturbation and prediction of the counterfactual question (i.e., Abduction). The second stage involves perturbing the feature subset of each sample that differs from their factual perturbation (i.e., Action) and then updating the predictions. 
Finally, we infer the PNS based on the proportion of changes and remaining unchanged in the prediction.
However, the conditional distribution in the first stage may be complex, we thus propose to use Sampling-Importance-Resampling~\cite{db1988using} on the observed inputs to approximate it. 
Further, we combine FANS and a gradient-based optimization method to extract the subset with the largest PNS.
We demonstrate FANS outperforms existing FAMs on six benchmarks of image and graph data.

\section{Causal Model for Feature Attribution}
In this section, we develop a principled framework to model the perturbation test in feature attribution, and describe the motivation of our proposed FANS approach that will be formally introduced in Section \ref{sec:formalize}. The essential notations are presented in Table \ref{tab:notation} of the Appendix.

\textbf{Structural Causal Model (SCM) for standard perturbation test.} In a SCM, all dependencies are generated by functions that compute variables from other variables. In this paper, we take SCM as our starting point and try to develop everything from there. Figure \ref{fig:process} illustrates the causal diagram of the standard perturbation test in feature attribution. Let $\mathbf{X}$ denote the $d$-dimensional input variable and $\mathbf{S}$ denotes a subset of dimensions $\{1, 2, ..., d\}$ of the input, with $\bar{\mathbf{S}}$ denoting its complement. We first observe the causal relationships $\mathbf{S} \to \tilde{\mathbf{X}}$ and $\mathbf{X} \to \tilde{\mathbf{X}}$. In this case, we sample around the target input $\mathbf{x}^t$ with the aim of taking into account the local behavior of the black box model~\cite{dhurandhar2022right}. 
Mathematically, let $b$ denote the local boundary around $\mathbf{x}^t$ and we define the \emph{neighborhood} of $\mathbf{x}^t$ as the distribution of $\mathbf{X}$ with the condition that the $L_p$ norm of $\mathbf{X}$ on $\mathbf{S}$ with respect to the value of $\mathbf{x}^t$ is not greater than $b$. To avoid confusion, we represent the corresponding variable as $\tilde{\mathbf{X}}$, as follows:
\begin{equation}\label{equ:condition}
\small
    \tilde{\mathbf{X}} \sim P(\mathbf{X} ~|~ \|\mathbf{X}_{\mathbf{S}} - \mathbf{x}^t_{\mathbf{S}}\|_p \leq b).
\end{equation}
Thus, the question is how to approximate the local boundary $b$ accurately. Our main idea is to define the boundary $b$ that perturbing samples within that boundary has a high probability of being Necessity and Sufficiency cause for the change in predictions. The formal definition of boundary $b$ will be introduced in Definition \ref{def:nsa}.
%
Note that the distribution in Eq. \ref{equ:condition} can be empirically approximated by sampling from the distribution of input given a value of $\mathbf{S}$ and selecting samples that meet the condition $\|\mathbf{X}_\mathbf{S} - \mathbf{x}^t_\mathbf{S}\|_p \leq b$.
We then observe the causal relationships $\tilde{\mathbf{X}} \to \mathbf{Y}$ and $\mathbf{S} \to \mathbf{Y}$. In this case, we perturb the features of $\tilde{\mathbf{X}}$ on $\mathbf{S}$ by a perturbation function $g$, and then feed the perturbed $\tilde{\mathbf{X}}$ into the model to generate a new prediction $\mathbf{Y}$, which is denoted by
\begin{equation}\label{equ:xs2y}
    \mathbf{Y} = f(g(\tilde{\mathbf{X}}, \mathbf{S}, \mathbf{x}')), 
\end{equation}
where we choose the most commonly used settings for baseline $\mathbf{x}'$ based on the data type and these details can be found in Appendix \ref{sec:imp}. Given a $d$-dimensional Bernoulli distribution with a  hyperparameter $\varphi$ (fixed to $0.5$ in practice) and an all-one vector $\bm{1}$, we implement the function $g$ to randomly perturb the features of $\tilde{\mathbf{X}}$ on $\mathbf{S}$ as follows.  
\begin{equation}\label{equ:perturb}
\small
         g(\tilde{\mathbf{X}}, \mathbf{S}, \mathbf{x}')  = (\bm{1} \!-\! \mathbf{m}_{\mathbf{S}}) \circ \tilde{\mathbf{X}}_{\mathbf{S}}  +   \mathbf{m}_{\mathbf{S}} \circ \mathbf{x}'_{\mathbf{S}},~\mathbf{m} \!\sim\! \text{Bern}(\bm{\phi}),
\end{equation}
where $\circ$ represents element-wise multiplication.
The reason for the Bernoulli distribution introduced in Eq.~\ref{equ:perturb} is that in the process of quantifying the model's predictions by the impact of perturbing the dimensions specified by $\mathbf{S}$, 
the dimensions that can influence the model predictions may be $\mathbf{S}$ or a subset of $\mathbf{S}$. Therefore, we sample selected dimensions of $\mathbf{S}$ to obtain a more comprehensive quantification.

\textbf{Goal of standard perturbation test.} 
A FAM using a standard perturbation test aims to quantify the importance of a feature subset corresponding to $\mathbf{s}$ (denoted by $a_{\mathbf{s}}$) which comprises two steps: (1) evaluate two model predictions of $\mathbf{Y}$ under different $\mathbf{S}$ values ($\mathbf{s}$ and $\mathbf{s}'$) through the causal model mentioned above, where $\mathbf{s}'$ used for reference is generally an empty set (i.e., no perturbation); (2) compute the importance $a_{\mathbf{s}}$ by comparing the difference in predictions. Formally,  $a_{\mathbf{s}}$  is given by a function $\phi: \mathcal{Y} \times \mathcal{Y} \to \mathcal{A}$ that maps two predictions in $\mathcal{Y}$ to a scalar in $\mathcal{A} \subseteq \mathbb R$.

\textbf{Remark 1.}
Most of the perturbation tests designed by various FAMs are compatible with our introduced causal model, contingent on their respective implementations of $\tilde{\mathbf{X}}$, score function $\phi$ and $\mathbf{s}'$. For example, Shapley Values~\cite{shapley1953value} fixes the variable $\tilde{\mathbf{X}}$ to be the target input $\mathbf{x}^t$, defines $\phi$ as Shapley Value, and sets $\mathbf{s} = \mathbf{s}' \cup x^t_i$.

\textbf{Remark 2.}
Our causal model differs from the standard causal generation process in two ways. First, variable $\mathbf{Y}$ is the model's prediction rather than a property of the real-world data. Second, by intervening on a variable in our process, the outcome of its corresponding children's variables is available.

\section{Feature Attribution as a Problem of PNS Measurement}\label{sec:formalize}
\begin{figure*}[t!]
	\centering
	\includegraphics[width=2\columnwidth]{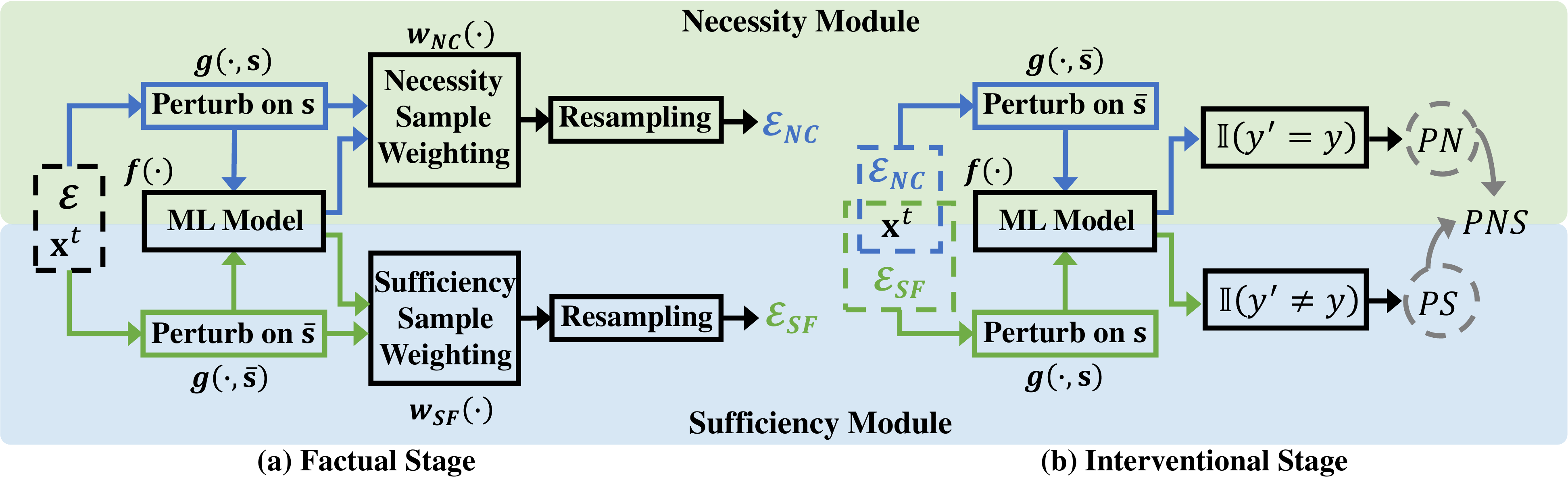} 
 \vspace*{-1.5mm}
	\caption{Architecture of FANS, which takes the sample $\mathbf{x}^t$ to be explained and the samples $\mathcal{E}\overset{\text{iid}}{\sim} P(\mathbf{X})$ as inputs, throughout the necessity and sufficiency modules to output PN and PS, and finally combine PN, PS into PNS. 
    Each module consists of two stages.
 1) Factual stage. 
 Generate samples $\mathcal{E}_{\text{NC}}$ and $\mathcal{E}_{\text{SF}}$ conditional on the fact that the model's predictions change or remain unchanged respectively after performing  perturbations on dimension subset $\mathbf{s}$ and $\bar{\mathbf{s}}$.
 2) Intervention stage. Apply perturbations different from the facts to $\mathcal{E}_{\text{NC}}$ and $\mathcal{E}_{\text{SF}}$, and count the proportion of changes and remaining unchanged by comparing the perturbed prediction $y'$ and the original prediction $y$. 
 }
\label{fig:model}
 \vspace*{-2mm}
\end{figure*}
In Example~\ref{exam:classifier}, we demonstrated that the feature attribution method using standard perturbation test as mentioned in the previous section, is limited in its ability to distinguish the importance of features when their changes in prediction are similar. To address this issue, our key finding is that similar changes in prediction can be caused by different causes. 
Building upon this finding, we now formally introduce a probability of causation, namely PNS, which quantifies the joint probability of two counterfactual questions about a feature subset. 
These counterfactual questions relate to two key events in the process of $\tilde{\mathbf{X}}, \mathbf{S} \to \mathbf{ Y}$ during the perturbation test. 
Specifically, consider an input $\mathbf{x}$, a subset of input dimensions $\mathbf{s}$ and let $\mathbf y$ denote the value of $\mathbf Y$ after feeding $\mathbf{x}$ and $\mathbf{s}$ into the causal model (Eqs.\ref{equ:xs2y}-\ref{equ:perturb}). 
We denote $A_{\mathbf{s},b}$ as an event of \emph{perturbation on the subset $\mathbf{s}$ of an input $\mathbf{x}$ sampled from $P(\mathbf{X} ~|~ \|\mathbf{X}_{\mathbf{S}} - \mathbf{x}^t_{\mathbf{S}}\|_p \leq b)$ (Eq.~\ref{equ:condition}), represented by the perturbation $g(\mathbf{x}, \mathbf{s}, \mathbf{x}')$ (Eq.~\ref{equ:perturb})}. 
We use $\bar{A}_{\mathbf{s},b}$ to denote the event of $A_{\mathbf{s},b}$ not occurring, represented by $\bar{A}_{\mathbf{s},b}$$=$$\vee^{\infty}_{i=1} A_{\mathbf{s}'_i,b'_i}$, where $\mathbf{s}'_i \ne \mathbf{s}, b'_i \ne b$. Since the number of possible pairs $(\mathbf{s}'_i, b'_i)$is large, we use the complement of $\mathbf{s}$ to approximate $\bar A_{\mathbf{s},b}$, thus $\bar A_{\mathbf{s},b} \approx A_{\bar s,b}$. 
We denote $B_{\mathbf{s},c}$ as an event representing \emph{the change in the original prediction $f(\mathbf{x})$ relative to $\mathbf{y}$, expressed as $|\mathbf{y}$$-$$f(\mathbf{x})|$$>$$c$}, where $c$ is used to determine if there is a significant difference between the perturbed and original predictions. The formal definition of $c$ will be introduced in Definition \ref{def:nsa}. Similarly, $\bar{B}_{\mathbf{s},c}$ is denoted as the event  $|\mathbf{y}$$-$$f(\mathbf{x})|$$\le$$c$. Based on these events,  we provide the following definitions of counterfactual probabilities~\cite{pearl2022probabilities}.
\begin{definition} (Probability of Necessity, PN)\label{equ:ddpn}
\begin{equation}
    PN = P(\bar{B}_{{\mathbf{s},c}_{\bar A_{\mathbf{s},b}}}|A_{\mathbf{s},b},B_{\mathbf{s},c}),
\end{equation}
PN represents the probability of the following counterfactual question: given that events $A_{\mathbf{s},b}$ and $B_{\mathbf{s},c}$ both occur initially, would $B_{\mathbf{s},c}$ not occur if  $A_{\mathbf{s},b}$ were changed from occurring to not occurring?
\end{definition}
\begin{definition}\label{equ:ddps}
(Probability of Sufficiency, PS)
\begin{equation}
    PS = P({B_{\mathbf{s},c}}_{A_{\mathbf{s},b}}|\bar A_{\mathbf{s},b}, \bar B_{\mathbf{s},c})
\end{equation}
PS represents the probability of the following counterfactual question:
given that events $A_{\mathbf{s},b}$ and $B_{\mathbf{s},c}$ both did not occur initially, would  $B_{\mathbf{s},c}$ occur if  $A_{\mathbf{s},b}$ were changed from not occurring to occurring?
\end{definition}
\begin{definition} (Probability of Necessity and Sufficiency, PNS)
    \begin{equation}
    \label{equ:pns_obj}
        PNS = PN \cdot P(A_{\mathbf{s},b}, B_{\mathbf{s},c}) + PS \cdot P(\bar A_{\mathbf{s},b}, \bar B_{\mathbf{s},c}).
    \end{equation}
PNS is the weighted sum of PN and PS, each multiplied by the probability of its corresponding condition. PNS measures the probability that $A_{\mathbf{s},b}$ is a necessary and sufficient cause for  $B_{\mathbf{s},c}$ (as the ``if and only if''  of the relationship).
\end{definition}

When evaluating the importance $w_{\mathbf{s}}$ of the input $\mathbf{x}^t$ on the dimension subset $\mathbf{s}$, we want $w_{\mathbf{s}}$ to reflect the probability of causation, then this objective can be formalized as follows: find a neighborhood for the target input $\mathbf x^t$ such that, compared to other neighborhoods of $\mathbf{x}^t$, perturbing samples within this neighborhood on $\mathbf{s}$ have the highest probability of being necessary and sufficient cause for the prediction change, and we take this probability as $w_{\mathbf{s}}$. Mathematically, the objective is to find the boundary $b$ and the threshold $c$ for events $A_{\mathbf{s}, b}$ and $B_{\mathbf{s}, c}$ that maximize the PNS, as illustrated in Definition \ref{def:nsa}, with an illustrative example provided. 
\begin{definition}\label{def:nsa} (Necessary and Sufficient Attribution) 
Necessary and Sufficient Attribution of the input $\mathbf{x}^t$ on the dimension subset $\mathbf{s}$ is defined as
\begin{equation}\label{equ:wi}
\small
    w_{\mathbf{s}} := \max_{b,c} PN \cdot P(A_{\mathbf{s},b}, B_{\mathbf{s},c})   +   PS \cdot P(\bar A_{\mathbf{s},b}, \bar  B_{\mathbf{s},c}). 
\end{equation}
\end{definition}
\textbf{Example 2.} Consider a $1$-dimensional model, $y$$=$$\sigma(\mathbf x^2)$, with a target sample $\mathbf x^t$$=$$0.2$, a baseline $\mathbf x'$$=$$0$, $\mathbf s$$=$$\{1\}$, $b$$=$$0.01$, and $c$$=$$0.1$. Since $\bar{\mathbf s}$ is empty, the domain of definition for $P(\tilde{\mathbf X} | \bar A_{\mathbf s,b}, \bar B_{\mathbf s,c})$ is empty. Thus, we focus  on calculating PN and $P(A_{\mathbf s,b}, B_{\mathbf s,c})$. 
Consider no perturbation will be performed when calculating PN since $\bar{\mathbf s}$ is empty. Consequently, PN equals 1. $P(A_{\mathbf s,b}, B_{\mathbf s,c})$ can be estimated through the empirical distribution of the dataset. Overall, the importance score of $\mathbf x^t$ is equal to $P(A_{\mathbf s,b}, B_{\mathbf s,c})$.


Therefore, to calculate the target score as shown in Eq.~\ref{equ:wi}, it can be broken down into the calculation of the boundary $b$, threshold $c$ and four probabilities: $P(A_{\mathbf{s},b}, B_{\mathbf{s},c})$ and $P(\bar A_{\mathbf{s},b}, \bar  B_{\mathbf{s},c})$ are the observation probabilities, while $P(\bar{B}_{{\mathbf{s},c}_{\bar A_{\mathbf{s},b}}} | A_{\mathbf{s},b}, B_{\mathbf{s},c})$ and $P(B_{{\mathbf{s},c}_{A_{\mathbf{s},b}}} | \bar A_{\mathbf{s},b}, \bar  B_{\mathbf{s},c})$ are the counterfactual probabilities PN and PS, respectively. 
According to the theory of counterfactual reasoning, PN and PS can be converted into the following equations for calculation (see Appendix \ref{sec:derivation} for detailed derivation).
\begin{equation}\label{equ:pn_def}
\small
\begin{aligned}
&P(\bar{B}_{{\mathbf{s},c}_{\bar{A}_{\mathbf{s},b}}} | A_{\mathbf{s},b}, B_{\mathbf{s},c})\\
    &= \mathbb E_{\mathbf{x} \sim P(\tilde{\mathbf{X}}|A_{\mathbf{s},b}, B_{\mathbf{s},c})} [P(|f(g(\mathbf{x}, \bar{\mathbf{s}}, \mathbf{x}')) - f(\mathbf{x})| \leq\! c)_{~\!\text{do}(\tilde{\mathbf{X}}=\mathbf{x})}],
\end{aligned}
\end{equation}
\begin{equation}\label{equ:ps_def}
\small
\begin{aligned}
    &P(B_{{\mathbf{s},c}_{A_{\mathbf{s},b}}} | \bar{A}_{\mathbf{s},b}, \bar  B_{\mathbf{s},c})\\
    &= \mathbb E_{\mathbf{x} \sim P(\tilde{\mathbf{X}}|\bar{A}_{\mathbf{s},b}, \bar{B}_{\mathbf{s},c})} [P(|f(g(\mathbf{x}, {\mathbf{s}}, \mathbf{x}')) - f(\mathbf{x})| >\! c)_{~\!\text{do}(\tilde{\mathbf{X}}=\mathbf{x})}],
\end{aligned}
\end{equation}
where $\text{do}(\tilde{\mathbf{X}}$$=$$\mathbf{x})$ denotes an intervention on  $\tilde{\mathbf{X}}$ that fixes the value as $\mathbf{x}$, i.e., perform the \emph{do}-calculus~\cite{2000Causality}. To compute the boundary $b$,  threshold $c$, and the probabilities $P(\bar{B}_{{\mathbf{s},c}_{\bar A_{\mathbf{s},b}}} | A_{\mathbf{s},b}, B_{\mathbf{s},c})$ and $P(B_{{\mathbf{s},c}_{A_{\mathbf{s},b}}} | \bar A_{\mathbf{s},b}, \bar  B_{\mathbf{s},c})$, we will propose a heuristic strategy for $b,c$ and a novel dual-stage perturbation test which implements the counterfactual reasoning paradigm, as introduced in Section \ref{sec:implementation}.
\section{Necessary and Sufficient Attribution \\ via Dual-stage Perturbation Test}\label{sec:implementation}
\subsection{Overview}\label{sec:method_overview}
By bridging the gap between PNS and feature attribution, we propose a method called Feature Attribution with Necessity and Sufficiency (FANS) to evaluate the Necessary and Sufficient Attribution (Definition \ref{def:nsa}) of each feature subset of the target input, as illustrated in Figure \ref{fig:model}. 
In particular, we first develop a novel dual-stage (factual and interventional) perturbation test that allows us to estimate counterfactual probabilities of events $A_{\mathbf{s},b}, B_{\mathbf{s},c}$. Next, we implement two dual-stage perturbation tests: one for PN and the other for PS, serving as the necessity and sufficiency modules of FANS, respectively. 
Sequentially, we combine the outputs of these modules to derive PNS. Finally, we propose a heuristic strategy to estimate the boundary $b$ and threshold $c$ that can maximize the PNS of $A_{\mathbf{s},b}, B_{\mathbf{s},c}$. Since the conditional distributions involved in the factual stage can be complex, FANS employs the Sampling-Importance-Resampling (SIR) method to approximate these distributions using observed samples. 
Additionally,  we  combine FANS with a gradient-based optimization to extract the feature subset with the highest Necessary
and Sufficient Attribution.

\subsection{Dual-stage Perturbation Test}
We design two different dual-stage perturbation tests to estimate the PN (Eq.~\ref{equ:pn_def}) and PS (Eq.~\ref{equ:ps_def}) by following the Abduction-Action-Prediction counterfactual reasoning paradigm \cite{pearl2013structural}. 
Abduction involves setting conditions for specific variables (corresponds to $\tilde{\mathbf{X}}$ in our context) based on  observations (corresponds to prediction $\mathbf{Y}$ given the dimensions $\mathbf{S}$). 
Action involves intervening in the values of certain variables (corresponds to $\mathbf{S}, \tilde{\mathbf{X}}$). Prediction involves computing the resulting conditional predictive distribution. 
This corresponds to the conditional distribution for $\mathbf{Y}$ given the intervention value of $\mathbf{S}$ and the factual values of $\tilde{\mathbf{X}}$.

We first present the dual-stage perturbation test for PS estimation.
\textbf{1) (Factual stage)} Draw inputs from a distribution conditional on the fact that predictions remain unchanged after using the perturbation function (Eq.~\ref{equ:perturb}) to perturb the features of $\tilde{\mathbf{X}}$ on $\overline{\mathbf{s}}$ (i.e., events $\bar{A}_{\mathbf{s},b}$ and $\bar{B}_{\mathbf{s},c}$ both occur).
\textbf{2) (Interventional stage)} Remove all arrows pointing to $\tilde{\mathbf{\mathbf{X}}}$ in Figure \ref{fig:process}. For each collected sample, use the perturbation function in Eq.~\ref{equ:perturb} to randomly perturb on $\mathbf{s}$ (i.e., event $A_{\mathbf{s},b}$ occur) and calculate the proportion of prediction changes (i.e., event $B_{\mathbf{s},c}$ occur). Finally, average the proportion of prediction changes for each input. Similarly, the dual-stage perturbation test for PN estimation is constructed by exchanging the perturbations and prediction events involved in the above two stages respectively.

\subsection{Sufficiency Module}
This module provides a concrete implementation of the dual-stage perturbation test for PS estimation.

\subsubsection{Factual Stage}
To draw samples from $P(\tilde{\mathbf{X}}|\bar{A}_{\mathbf{s},b},\bar{B}_{\mathbf{s},c})$ in Eq.~\ref{equ:ps_def}, we propose applying SIR on the observed sample set $\mathcal{E}$ that follows $P(\mathbf{X})$ to approximate the required conditional distribution.
First, given the subset of dimensions $\bar{\mathbf{s}}$, let $r$ denote a normalization constant and the resampling weight for $\mathbf{x}$ sampled from $P(\mathbf{X})$ is as follows (see Appendix \ref{sec:sir} for derivation). 
\begin{equation}\label{equ:hard_weight_pn}
\small
\begin{aligned}
    \!w_{\text{SF}}(\mathbf{x})\!=\!\left\{\begin{array}{lr}
\!\!\!r \!\cdot\! P(|f(g(\mathbf{x}, \bar{\mathbf{s}}, \mathbf{x}')) \!-\! f(\mathbf{x})| \!\leq c|~\mathbf{x}), &\!\!\!\!\!\!\!\text{ if } \|\mathbf{x}_{\bar{\mathbf{s}}}\!-\!\mathbf{x}^t_{\bar{\mathbf{s}}}\|_p\leq b \\
\!\!\!0, & \text { otherwise.}
\end{array}\right.
\end{aligned}
\end{equation}
Hence, the resampling weight is determined by two conditions: $\|\mathbf{x}_{\bar{\mathbf{s}}}-\mathbf{x}^t_{\bar{\mathbf{s}}}\|_p\leq b$ and $|f(g(\mathbf{x}, \bar{\mathbf{s}}, \mathbf{x}')) - f(\mathbf{x})| \leq c$. 
To enhance the smoothness of $w_{\text{SF}}(\cdot)$, we introduce a  Gaussian kernel function to soften the above two 
conditions and take their product as an approximation of $w_{\text{SF}}(\mathbf{x})$, as follows.
\begin{equation}\label{equ:soft_weight_ps}
\small
    \tilde{w}_{\text{SF}}(\mathbf{x}) := \exp\big(\frac{\|\mathbf{x}_{\bar{\mathbf{s}}} - \mathbf{x}^t_{\bar{\mathbf{s}}}\|^2_p}{-2b^2}\big) \cdot \exp\big(\frac{|f(g(\mathbf{x}, \bar{\mathbf{s}}, \mathbf{x}')) \!-\! f(\mathbf{x})|^2}{-2c^2}\big),
\end{equation}
where the methods for estimating $b$ and $c$ will be introduced in Section \ref{sec:bc}.  
We then feed observed samples and their weight according to Eq.~\ref{equ:soft_weight_ps} into the SIR algorithm to obtain a sample set that approximates the conditional distribution $P(\tilde{\mathbf{X}}|\bar{A}_{\mathbf{s},b}, \bar{B}_{\mathbf{s},c})$ in PS (Eq.~\ref{equ:ps_def}).
\subsubsection{Intervention Stage}
Let $\mathcal{E}_{\text{SF}}$ be a sample set of  $P(\tilde{\mathbf{X}}|\bar{A}_{\mathbf{s},b}, \bar{B}_{\mathbf{s},c})$ obtained through SIR.
Then, PS (Eq.~\ref{equ:ps_def}) can be estimated as follows:
\begin{equation}\label{equ:ps_estimate}
\small
\begin{aligned}
        \hat{\text{PS}} =\frac{1}{|\mathcal{E}_{\text{SF}}|\cdot t_{\text{SF}}} \sum_{\mathbf{x} \in \mathcal{E}_{\text{SF}}}\sum^{t_{\text{SF}}}_{j=1} \mathbb I(|f(g({\mathbf{x}}, \mathbf{s}, \mathbf{x}')) - f({\mathbf{x}})| > c),
\end{aligned}
\end{equation}
where $t_{\text{SF}}$ is a hyperparameter representing the number of times a random perturbation is applied to $\mathbf{x}$ through the perturbation function $g$ (Eq. \ref{equ:perturb}).

\subsection{Necessity Module}
The implementation of the necessity module is similar to the sufficiency module, differing only in the opposite perturbation and prediction events $A_{\mathbf{s},b}, B_{\mathbf{s},c}$ in each stage.
Thus, in the factual stage, the resampling weight for the observed input $\mathbf{x}$ sampled from $P(\mathbf{X})$, 
which is used to  generate the sample set approximating the conditional distribution $P(\tilde{\mathbf{X}}|A_{\mathbf{s},b}, B_{\mathbf{s},c})$ in Eq.~\ref{equ:pn_def}, is given by
\begin{equation}\label{equ:soft_weight_pn}
\small
    \!\!\tilde{w}_{\text{NC}}(\mathbf{x}) \!:=\! \exp\big(\frac{\|\mathbf{x}_{\mathbf{s}} \!-\! \mathbf{x}^t_{\mathbf{s}}\|^2_p}{-2b^2}\big) \cdot \big(1\!-\exp\big(\frac{|f(g(\mathbf{x}, \mathbf{s}, \mathbf{x}')) \!-\! f(\mathbf{x})|^2}{-2c^2}\big)\big).
\end{equation}
Further, let $\mathcal{E}_{\text{NC}}$ be a sample set of $P(\tilde{\mathbf{X}}|A_{\mathbf{s},b}, B_{\mathbf{s},c})$ obtained through SIR. Then PN (Eq.~\ref{equ:pn_def}) can be estimated as follows:
\begin{equation}\label{equ:pn_estimate}
\small
        \hat{\text{PN}} = \frac{1}{|\mathcal{E}_{\text{NC}}| \cdot t_{\text{NC}}}\sum_{\mathbf{x} \in \mathcal{E}_{\text{NC}}}\sum^{t_{\text{NC}}}_{j=1} \mathbb I(|f(g(\mathbf{x}, \bar{\mathbf{s}}, \mathbf{x}')) - f(\mathbf{x})| \leq c).
\end{equation}
where the constant $t_{\text{NC}}$ denotes the number of perturbations.

\subsection{PNS Estimation}
As indicated by Eq.~\ref{equ:pns_obj}, PNS is equivalent to the weighted sum of PS and PN, with the weights being $P(A_{\mathbf{s},b}, B_{\mathbf{s},c})$ and $P(\bar A_{\mathbf{s},b}, \bar B_{\mathbf{s},c})$. The weights can be calculated as follows.

Consider  that $P(A_{\mathbf{s},b}, B_{\mathbf{s},c})$ (or $P(\bar A_{\mathbf{s},b}, \bar B_{\mathbf{s},c})$) can be expressed as the weighted sum of conditional probabilities $P (A_{\mathbf{s},b}, B_{\mathbf{s},c}|\mathbf{x})$ with respect to all possible $\mathbf{x}$, where the weight is $P(\mathbf{x})$. Thus, we can estimate $P(A_{\mathbf{s},b}, B_{\mathbf{s},c})$ and $P(\bar A_{\mathbf{s},b}, \bar B_{\mathbf{s},c})$ by summing their SIR-based weights of the observed samples in Eq.~\ref{equ:soft_weight_pn} and Eq.~\ref{equ:soft_weight_ps} respectively. In particular,
given the sample set $\mathcal{E}\overset{\text{iid}}{\sim} P(\mathbf{X})$,
\begin{equation}\label{equ:trade_off}
\small
    \!\!\hat{P}(A_{\mathbf{s},b}, B_{\mathbf{s},c})=\sum_{\mathbf{x}\in \mathcal{E}} w_{\text{NC}}(\mathbf{x}),\quad
    \hat{P}(\bar{A}_{\mathbf{s},b}, \bar{B}_{\mathbf{s},c}) = \sum_{\mathbf{x} \in \mathcal{E}}w_{\text{SF}}(\mathbf{x}).
\end{equation}

\subsection{Necessary and Sufficient Attribution Estimation}\label{sec:bc}
As illustrated in Definition \ref{def:nsa}, in order to estimate our proposed Necessary and Sufficient Attribution, it is necessary to determine the boundary $b$ and threshold $c$ of events $A_{\mathbf{s},b}$ and $B_{\mathbf{s},c}$ that maximize PNS. FANS employs the following heuristics to compute their values.


The value of $b$ is set to ensure that the neighborhood of $\mathbf{x}^t$ is both small and nearly uniformly dense. Thus, the value of $b$ is determined based on the Scott rule~\cite{scott1979optimal}, a bandwidth estimation method widely used in kernel density estimation, which is given by $b=1.06 \cdot|\mathcal{E}|^{\frac{1}{4+d}}$, where $\mathcal{E} \stackrel{\mathrm{iid}}{\sim} P(\mathbf{X})$ and $d$ is the dimension of the input. 
On the other hand, the strategy for determining the value of $c$  is as follows. First, we  use the normal distribution $N(\mathbf 0, \sigma^2 \cdot \mathbf I)$ to simulate low-intensity noise, where $\sigma$ is a small constant (set to 0.001 in practice), $\mathbf 0$ is the all-zero vector, and $\mathbf I$ is the identity matrix. Assuming the model's predictions for inputs with this noise remain unchanged, we define the maximum variance of these predictions as $c:=\max _{\mathbf{x} \in \mathcal{E}, \epsilon \in \mathcal{R}}|f(\mathbf{x})-f(\mathbf{x}+\epsilon)|$, where $\mathcal{E} \stackrel{\mathrm{iid}}{\sim} P(\mathbf{X}), \mathcal R \stackrel{\mathrm{iid}}{\sim} N(\mathbf 0, \sigma^2 \cdot \mathbf I)$. The sensitivity analysis of $b$ and $c$ are provided in Appendix \ref{sec:fine_tuning}.

\subsection{Extracting Feature Subset with the Highest Necessary and Sufficient Attribution}\label{sec:optimize}
In this section, we aim to efficiently extract the feature subset with the highest estimated Necessary and Sufficient Attribution in the target input. Specifically, we use stochastic gradient descent to maximize the estimated Necessary and Sufficient Attribution with respect to the dimension subset $\mathbf{s}$. Given the discrete nature of the dimension subset s, we first encode it as a binary vector containing $d$ dimensions, and then relax the entries in the discrete set $\{0, 1\}$ into the continuous range $[0, 1]$. Additionally, 
we use $\bm{1}$$-$$\mathbf{s}$ to represent the complement of $\mathbf{s}$, where $\bm{1}$ is an all-one vector. 
Further, we replace $\|\mathbf{x}_{\bar{\mathbf{s}}} - \mathbf{x}^t_{\bar{\mathbf{s}}}\|_p$ in Eq.~\ref{equ:soft_weight_ps} with $\|\mathbf{x} \circ (\bm{1} - \mathbf{s}) - \mathbf{x}^t \circ (\bm{1} - \mathbf{s}) \|_p$, where $\mathbf{s}$ is now a relaxed binary vector and $\circ$ represents element-wise multiplication. Similarly, we replace $(\bm{1} \!-\! \mathbf{m}_{\mathbf{S}}) \circ \tilde{\mathbf{X}}_{\mathbf{S}}  +   \mathbf{m}_{\mathbf{S}} \circ \mathbf{x}'_{\mathbf{S}}$ in Eq.~\ref{equ:perturb} with $(\bm{1} \!-\! \mathbf{m} ) \circ \tilde{\mathbf{X}} \circ \mathbf{s}  +   \mathbf{m}  \circ \mathbf{x}' \circ \mathbf{s}$.
In addition, the indicator function $\mathbb I(|f(\cdot) \!-\! f(\cdot)| > c)$ in Eq.~\ref{equ:ps_estimate} is not differentiable, so we replace it with the smooth function $1 \!- \exp{(-|f(\cdot) \!- \!f(\cdot)|)}$ to indicate whether the prediction changes. 
Similarly, $\mathbb I(|f(\cdot)  - f(\cdot)| \leq c) $ in Eq.~\ref{equ:pn_estimate} is replaced with $\exp(-|f(\cdot) - f(\cdot)|)$.

After optimizing with respect to $\mathbf{s}$ by maximizing Eq.~\ref{equ:wi}, 
the resultant $\mathbf{s}$ is used for the soft selection of the feature subset with the highest Necessary and Sufficient Attribution.

\section{Experiments}
\begin{table*}[htbp]
 \renewcommand{\arraystretch}{0.75}
  \centering
  \vspace*{-4.8mm}
  \caption{Performance on the image datasets. The best performance is marked in bold, and the second best is underlined. Symbols $\uparrow$ and $\downarrow$ respectively represent that larger and smaller metric values are better.}
  \begin{small}
  \resizebox{2\columnwidth}{!}{%
          \begin{tabular}{lccccrccccrcccc}
    \toprule
          & \multicolumn{4}{c}{MNIST}     &       & \multicolumn{4}{c}{Fashion-MNIST} &       & \multicolumn{4}{c}{CIFAR10} \\
\cmidrule{2-5}\cmidrule{7-10}\cmidrule{12-15}    \multicolumn{1}{c}{Method} & INF$\downarrow$  & \multicolumn{1}{l}{IR$\uparrow$} & SPA$\uparrow$  & MS$\downarrow$   &       & INF$\downarrow$  & \multicolumn{1}{l}{IR$\uparrow$} & SPA$\uparrow$  & MS$\downarrow$   &       & INF$\downarrow$  & \multicolumn{1}{l}{IR$\uparrow$} & SPA$\uparrow$  & MS$\downarrow$ \\
    \midrule
    Saliency & ${3.8 \times 10^4}$ & 64.3 & 0.658  & 0.623  &       & ${1.8 \times 10^6}$  & 25.5  & 0.558 & 0.753 &       & ${1.2 \times 10^8}$ & 54.1 & 0.492 & 0.736 \\
    IG &   \underline{${1.7 \times 10^3}$}    &   \underline{73.3}    &   \underline{0.918}    &   0.683    &       &   \underline{${1.7 \times 10^4}$}    &   \underline{60.8}    &   0.612    &    0.806   &       &  ${1.5 \times 10^5}$  &   \underline{63.5}    &  0.631    &  0.966 \\
    DeepLift & ${2.2 \times 10^3}$ & \underline{73.3}   & \underline{0.918} & 0.679 &       & ${8.2 \times 10^4}$  & 59.8  & 0.610 & 0.797 &       & ${1.9 \times 10^5}$     & 62.7 & 0.631 & 0.959 \\
    IDGI & ${2.0 \times 10^3}$ & 64.7 & 0.837  & 0.578  &       & ${2.6 \times 10^4}$  & 58.2  & 0.593 & 0.781 &       & \underline{${2.3 \times 10^4}$} & 19.5 & \underline{0.632} & 0.854 \\
    GradShap & ${2.2 \times 10^3}$ & \underline{73.3}   & \underline{0.918} & 0.673 &       & ${2.5 \times 10^4}$  & 59.2  & \underline{0.614} & 0.874 &       & ${2.3 \times 10^5}$ & 56.5 & 0.630 & 1.000 \\
    \midrule
    LIME & ${6.6 \times 10^5}$ & 67.9 & 0.808  & 0.899  &       & ${2.5 \times 10^8}$  & 28.6  & 0.533 & 0.884 &       & ${1.1 \times 10^9}$ & 2.1 & 0.512 & 1.032 \\
    Occlusion & ${5.7 \times 10^5}$   & 69.2 & 0.802 & \underline{0.538} &       & ${2.8 \times 10^7}$  & 58.4 & 0.505 & 0.660 &       & ${2.1 \times 10^8}$ & 51.4 & 0.507 & 0.857 \\
    FeatAblation & \underline{${1.7 \times 10^3}$} & 72.8  & 0.917 & 0.669 &       & ${5.5 \times 10^4}$ & 45.4  & 0.572 & 0.791 &       & ${1.1 \times 10^6}$ & 36.8 & 0.619 & 0.983 \\
    MP &${8.5\times10^5}$     &   70.3    &   0.421    &   0.904    &       &   ${2.9 \times 10^7}$    &   20.1    &  0.227     &   \textbf{0.453}   &       &     ${5.1 \times 10^8}$  &   16.3    & 0.476      & 0.887 \\
    CIMI  & ${1.7 \times 10^4}$ & 12.9  & 0.901 & 0.548 &       & ${2.6 \times 10^4}$   & 54.8 & 0.589 & 0.827 &       & ${6.0 \times 10^6}$ & 21.1 & 0.589 & \underline{0.615} \\
    \midrule
    FANS & $\mathbf{9.0 \times 10^2}$ & \textbf{74.5} & \textbf{0.924} & \textbf{0.463} &       & $\mathbf{{1.2 \times 10^4}}$ & \textbf{63.1}  & \textbf{0.630} & \underline{0.586} &       & $\mathbf{1.7 \times 10^4}$ & \textbf{63.6} & \textbf{0.634} & \textbf{0.578} \\
    \bottomrule
    \end{tabular}}%
  \end{small}
  \label{tab:performance_img}%
 \vspace*{-1mm}
\end{table*}%
\subsection{Experimental Setup}
1) \textbf{Datasets and models.} 
We utilized six public datasets with well-trained models: CIFAR10 \cite{krizhevsky2009learning} with ResNet18, MNIST     \cite{lecun1998gradient} and Fashion-MNIST \cite{xiao2017fashion} with LeNet5, as well as BACommunity, Pubmed, and Citeseer with GCN \cite{ying2019gnnexplainer,sen2008collective,kipf2016semi}. CIFAR10, MNIST, and Fashion-MNIST are image datasets. BACommunity, Pubmed, and Citeseer are graph datasets. Please see Appendix \ref{sec:dataset} for details.

2) \textbf{Baselines.} 
We selected the most popular or the most recent methods as the baseline. We then compared FANS with six feature-wise attribution methods: 
Saliency \cite{simonyan2013deep}, GuidedBP \cite{DBLP:journals/corr/SpringenbergDBR14}, IntegratedGrad(IG) \cite{sundararajan2017axiomatic}, GradShap \cite{lundberg2017unified}, DeepLift \cite{DBLP:conf/iclr/AnconaCO018}, and IDGI  \cite{yang2023idgi}, 
as well as twelve feature subset-wise attribution approaches: 
Occlusion \cite{zeiler2014visualizing}, 
LIME \cite{ribeiro2016should}, MeaningfulPerturb(MP) \cite{fong2017interpretable}, 
GNNExplainer \cite{ying2019gnnexplainer}, PGExplainer \cite{luo2020parameterized}, FeatAblation \cite{kokhlikyan2020captum},
ReFine \cite{wang2021towards}, 
CFGNExplainer \cite{lucic2022cf}, 
CF$^2$ \cite{tan2022learning}, 
NSEG~\cite{cai2022probability},
CIMI \cite{wu2023causality} 
and MixUpExplainer \cite{zhang2023mixupexplainer}. The data categories used to test each baseline are largely aligned with their literature. Except that 
we adjust CIMI to accommodate image data, and run Saliency, GuidedBP, and IG on both graph and image data. Please see Appendix \ref{sec:baselines} for details.

3) \textbf{Evaluation metrics.} 
We utilized six widely used metrics to evaluate the effectiveness of FANS: 
Infidelity(INF) \cite{yeh2019fidelity}, 
Iterative Removal Of Features(IR) \cite{DBLP:journals/corr/abs-2003-08747}, 
Fidelity$^+$(FID$^+$), 
Fidelity$^-$(FID$^-$) \cite{DBLP:conf/log/AmaraYZHZSBS022}, 
Max-Sensitivity(MS) \cite{yeh2019fidelity},
Sparseness(SPA) \cite{DBLP:conf/icml/ChalasaniC00J20},
and Recall@12 \cite{wang2021towards} metrics. 
INF, IR, FID$^+$, and FID$^-$ measure the extent to which the explanation follows the predicted behavior of the model (i.e., faithfulness), 
while MS measures the stability of the explanation when subjected to slight input perturbations (i.e., robustness) \cite{hedstrom2023quantus}. 
SPA measures the sparsity of the explanation by using the Gini Index.
Recall@$12$ measures the number of the top 12 most important features for a given attribution are features of the ``ground-truth explanation''. 
Remarkably, Recall@$12$ is only suitable for BACommunity since this dataset contains ``ground-truth explanation''. See Appendix \ref{sec:metric} for details. 

To test the explanation methods, we randomly selected 4,000 images from each image dataset (i.e., MNIST, Fashion-MNIST, and CIFAR10) and adopted the same configuration as the literature \cite{wang2021towards} on the graph datasets (i.e., BACommunity, Citeseer, and Pubmed). The results presented in Tables \ref{tab:performance_img} and \ref{tab:performance_g} are largely averages of the metric values for the different inputs, with the exception of FID$^+$ and FID$^-$, since these are metrics designed for the entire dataset rather than an individual sample. As each graph dataset we use contains only one graph, and FANS requires multiple samples for SIR, we create a graph-augmented dataset for each graph by randomly masking their adjacency matrix. See Appendix \ref{sec:imp} for more implementation details.
\subsection{Explanation Evaluation}
\subsubsection{Quantitative evaluation}
\textbf{Performance comparison.} We verify the experimental performance of FANS against the state-of-the-art baselines in Tables \ref{tab:performance_img} and \ref{tab:performance_g}. As shown in Table~\ref{tab:performance_img}, 
FANS significantly and consistently outperforms all the baselines in terms of faithfulness, sparsity, and robustness on different image datasets. In particular, we find that:
1) For the faithfulness metric INF, FANS outperforms the second-best approach by 47.06\%, 29.41\%, and 26.09\% on MNIST, Fashion-MNIST, and CIFAR10. For IR, FANS improves by 1.64\%, 3.78\%, and 0.16\%. These results indicate that extracting necessary and sufficient causes for input predictions can effectively capture key subsets of features that influence the model's predictive behavior.
2) For the sparsity metric SPA, FANS outperforms the next best method by 0.65\%, 2.61\%, and 0.32\% on the three image datasets. This validates that FANS's necessity module, which adaptively removes unimportant features by randomly masking them, outperforms other methods in terms of attribution sparsity. 
3) For the robustness metric MS, FANS outperforms the second-best method by 13.94\% and 6.02\% on the MNIST and CIFAR10 datasets, while achieving a somewhat lower but still competitive MS result on Fashion-MNIST. We find that FANS significantly outperforms gradient-based methods, probably due to the sensitivity of gradients to noisy feature values. FANS learns from a set of similar features obtained through SIR, which helps it to mitigate the effects of noise.

Similarly, in Table \ref{tab:performance_g}, FANS regularly surpasses all the baselines on the graph datasets. In particular, in terms of FID+, FANS improves by 4.99\% on BACommunity, 46.15\% on Citeseer, and 13.33\% on Pubmed relative to the baselines. It provides further modest improvements with respect to the sparsity metric SPA.
For another faithfulness metric FID-, FANS is the only method achieving optimal values (i.e., 0)  across all three datasets. FANS outperforms other approaches for Citeseer and Pubmed while remaining competitive on Recall@12 for the synthetic BACommunity dataset.  
\begin{table*}
\renewcommand{\arraystretch}{0.75}
    \centering
  \caption{Performance on the graph datasets.}
  \begin{small}
          \begin{tabular}{lccccrcccrccc}
    \toprule
          & \multicolumn{4}{c}{BACommunity} &       & \multicolumn{3}{c}{Citeseer} &       & \multicolumn{3}{c}{Pubmed} \\
\cmidrule{2-5}\cmidrule{7-9}\cmidrule{11-13}    \multicolumn{1}{c}{Method} & FID$^+$$\uparrow$ & FID$^-$$\downarrow$ & SPA$\uparrow$  & Recall@12$\uparrow$ &       & FID$^+$$\uparrow$ & FID$^-$$\downarrow$ & SPA$\uparrow$  &       & FID$^+$$\uparrow$ & FID$^-$$\downarrow$ & SPA$\uparrow$ \\
    \midrule
    Saliency & \underline{0.6667} & 0.1000   & 0.8899 & 0.5141 &       & 0.1800  & \underline{0.0400}  & 0.7476 &       & 0.0250 & \underline{0.0250} & \underline{0.9670} \\
    GuidedBP & 0.4833 & 0.1000   & 0.9137 & 0.3822 &       & 0.0607  & 0.1207  & \underline{0.8410} &       & 0.0000     & \underline{0.0250} & 0.9629 \\
    IG & \underline{0.6667} & 0.1167 & \underline{0.9200}  & 0.5204 &       & \underline{0.5200}  & \underline{0.0400}  & 0.8380 &       & 0.0250 & \textbf{0.0000} & 0.9636 \\
    \midrule
    GNNExplainer & 0.4000   & 0.0667 & 0.6432 & 0.4210 &       & 0.1821  & \textbf{0.0000} & 0.3760 &       & 0.0750 & \textbf{0.0000} & 0.6058 \\
    PGExplainer & 0.4667 & \underline{0.0500}  & 0.8092 & \underline{0.6025} &       & 0.4292 & 0.2800  & 0.1435 &       & 0.0258 & \underline{0.0250} & 0.01380 \\
    ReFine &   0.4833    &   0.1500    &   0.4728    &   0.3208    &       &   0.2600    &   0.2600    &   0.4623    &       &   0.1055    &   0.1000    &  0.5890\\
    CFGNExplainer & 0.5167 & 0.1500  & 0.7199 & 0.1987 &       & 0.5011   & \textbf{0.0000} & 0.5454 &       & \underline{0.3751} & \textbf{0.0000} & 0.7048 \\
    CF$^2$ &   0.4667    &   0.0667    &   0.6216    &  \textbf{0.6189}     &       &   0.2200    &  \textbf{0.0000}     &  0.3706     &       &  0.1250     &    \textbf{0.0000}   & 0.5991 \\
    NSEG &   0.5167    &   0.1000    &   0.6258    &  0.4806     &       &   0.1600    &  \textbf{0.0000}     &  0.3269     &       &    0.0750   &    \textbf{0.0000}  & 0.5757 \\
    MixUpExplainer &   0.2333    &   0.1500    &   0.2064    &  0.3708     &       &   0.1200    &  0.0800     &  0.2963     &       &  0.0250     &    0.0750   & 0.1041 \\
    \midrule
    FANS & \textbf{0.7000} & \textbf{0.0000} & \textbf{0.9311} & 0.5819 &       & \textbf{0.7600} & \textbf{0.0000}  & \textbf{0.8619} &       & \textbf{0.4251} & \textbf{0.0000} & \textbf{0.9755} \\
    \bottomrule
    \end{tabular}%
  \end{small}
  \label{tab:performance_g}%
\end{table*}
\begin{figure}[t!]
	\centering
	\includegraphics[width=1\columnwidth]{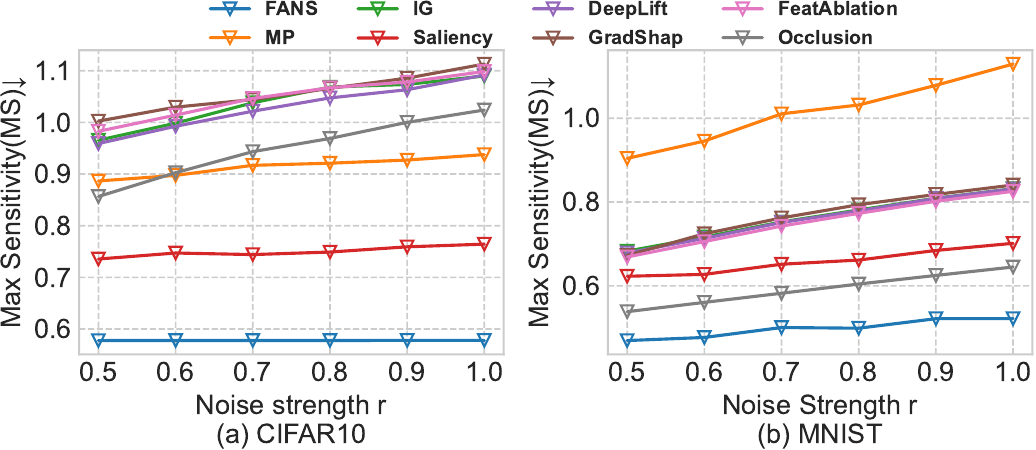} 
 \vspace*{-7.4mm}  
	\caption{Performance on robustness comparison in image datasets CIFAR10 and MNIST under different strengths of noise $r$. }
 \vspace*{-4.8mm}
        \label{fig:robustnes}
\end{figure}

\textbf{Robustness analysis.} Next, we investigate the robustness of FANS and the baseline methods by examining the effect of introducing noise. 
Following the literature~\cite{yeh2019fidelity}, we designed a robustness experiment as follows. Given various values of $r$, for $r\in [0, 1]$, we add noise to each pixel of input image $\mathbf{x}_i$ from the interval [$-r$, $r$]. We then evaluate the difference between the explanations of each perturbed image and its original explanation via the Max Sensitivity (MS) metric. As shown in Fig.\ref{fig:robustnes}(a)(b), as expected, all methods demonstrate a gradual deterioration in performance with increased noise strength. However, the slope of this deterioration is considerably lower using FANS, which demonstrates far superior robustness on both examined datasets.  
This can be explained by the fact that FANS learns from multiple samples which are similar to the target input through SIR, rather than a single sample, which renders it more robust to perturbation noise.

\subsubsection{Qualitative evaluation}
\begin{figure}[t!]
	\centering
	\includegraphics[width=0.90\columnwidth]{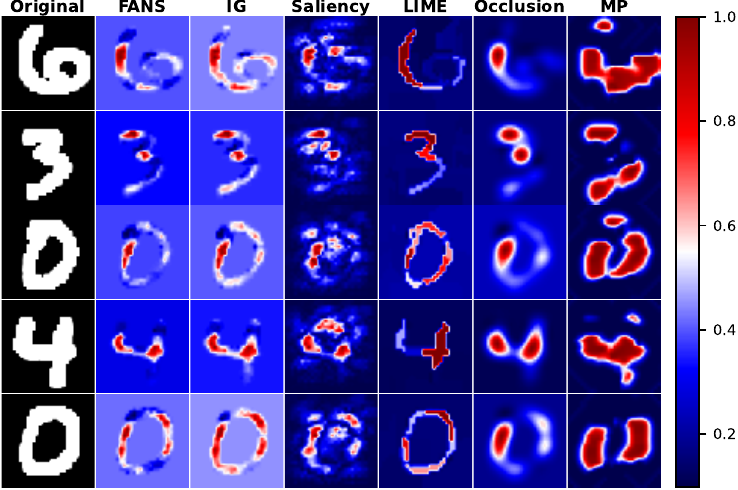} 
 \vspace*{-4mm}
	\caption{Attributions visualization on the MNIST dataset.}
  \vspace*{-5mm}
 \label{fig:heatmap}
\end{figure}
In Figure \ref{fig:heatmap}, we visualize the attributions obtained for FANS and other baselines on five samples from the MNIST dataset. 
We find that the attributions of FANS and IG are more sparse than other methods. Compared to IG, FANS improves the contrast between the scores of important and unimportant features. This is visually demonstrated by a higher number of blue pixels in FANS than in IG, which is expected since FANS's PNS is used to attribute the sparsest and the most influential feature subset for the model's predictions. 

\subsection{Ablation Study}
\begin{figure}
    \centering
    \subfigure[Citeseer]{
        \includegraphics[width=0.45\columnwidth]{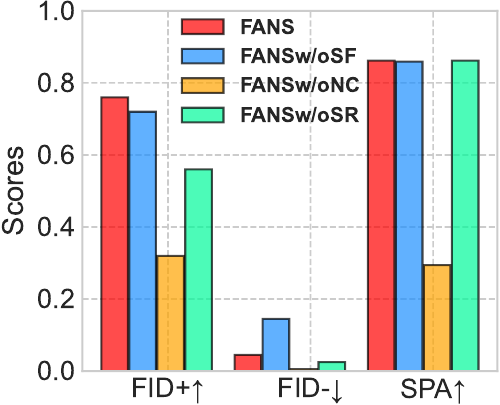}
        \label{fig:ns_abdu_citeseer}
    }  
    \subfigure[Pubmed]{
    \includegraphics[width=0.45\columnwidth]{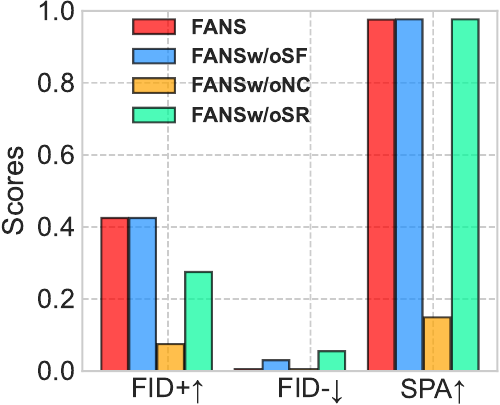}
    \label{fig:ns_abdu_pubmed}
    }
    \vspace*{-3.8mm}
    \caption{Ablation study of sufficiency (SF) module, necessity (NC)  module, and SIR-based Sampling (SR) on graph datasets Citeseer and Pubmed.}
    \vspace*{-3mm}
    \label{fig:ablation}
\end{figure}
The core components of FANS are the necessity, sufficiency, and SIR-based Sampling modules. In order to validate the importance of each component and better understand their relationship to different explainability metrics, we ablate our method by iteratively removing each component. We refer to FANS without the sufficiency module as FANSw/oSF, FANS without the necessity module as FANSw/oNC, and FANS without SIR-based Sampling as FANSw/oSR. We visualize the effects of this ablation on performance relative to the full algorithm in Figure \ref{fig:ablation} for the Citeseer and Pubmed datasets. 
In Figure \ref{fig:ablation}(a)(b), we first remark that the full FANS algorithm is the only version of the method to perform uniformly well across both datasets and all metrics. 
When comparing with FANSw/oSF,  we notice a slight decrease in the faithfulness metric FID+ and a significant decrease in FID-, while the sparsity metric remains stable, where FID- and FID+  evaluate the proportions of predictions remain unchanged and change when features with high attribution scores are retained and removed, respectively.
Conversely, FANSw/oNC exhibits an opposite trend to the above three metrics of FANSw/oSF, with FID+ becoming significantly worse, FID- remaining stable, and SPA becoming worse. 
This result is expected since the sufficiency and necessity modules ensure that the feature subset contains all features responsible for the prediction and removes those that are not respectively, which result in a more dense or sparse optimal feature subset. 
Thus, the optimal subset estimated by FANSw/oSF may only contain a subset of features that are responsible for prediction. Removing these features may lead to prediction change (high FID+), while retaining these features may also lead to prediction change (high FID-) since the important features are incomplete. 
Similarly, the optimal subset estimated by FANSw/oNC may contain redundancy. Retaining these features may keep the prediction unchanged (low FID-) 
while removing these features the change in prediction is uncertain (low FID+) due to the redundancy.

Next, we compared FANS with FANSw/oSR. As shown in Figure \ref{fig:ablation}(a)(b), the sparsity metric SPA of FANSw/oSR remains stable because of FANS's necessity module. However, both faithfulness metrics FID+ and FID- decrease. This may be because FANSw/oSR lacks information regarding samples that are similar to the target input when learning the optimal subset of features, which may lead to FANS being affected by unimportant features of the target input. 
\subsection{Convergence Analysis}
\begin{figure}[t]
\centering
\includegraphics[width=\columnwidth]{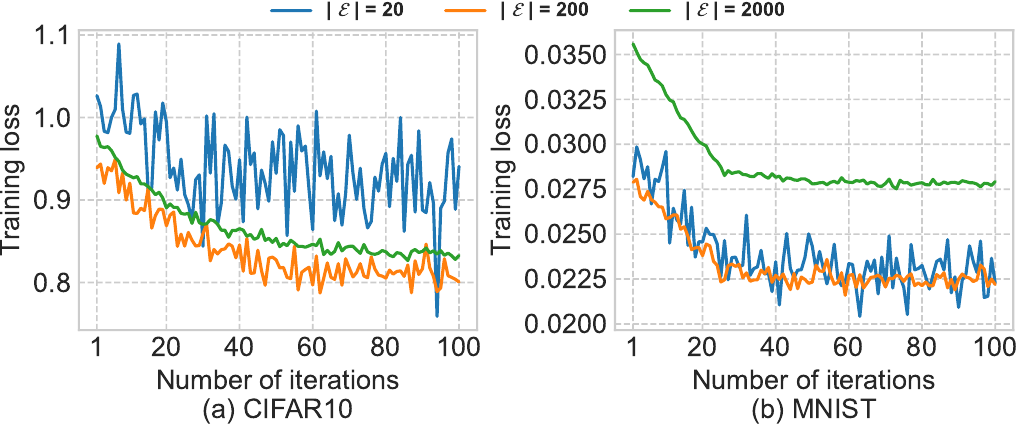}
 \vspace*{-7.5mm}
\caption{Convergence of FANS on datasets CIFAR10 and MNIST with sample sizes $|\mathcal{E}|=20, |\mathcal{E}|=200$ and $|\mathcal{E}|=2,000$.}
\label{fig:convergence}
\vspace*{-2mm}
\end{figure}

We corroborate the convergence analysis of FANS under different sizes of $\mathcal{E}$ which are used for SIR-based resampling and show our FANS approach a stationary point. Figure \ref{fig:convergence} shows the learning process of FANS with sample size $|\mathcal{E}|=20, |\mathcal{E}|=200$ and $|\mathcal{E}|=2,000$. These values correspond to varying degrees of sparsity within the sample set $\mathcal{E}$.  The training losses decrease with the number of iterations, leading to a convergent result in all cases. As shown in Figure \ref{fig:convergence}~(a), we find that the training loss falls rapidly within the first 35 iterations, and begins to converge gradually in about 60 epochs. Figure \ref{fig:convergence}~(b) shows that the model achieves earlier convergence within the first 30 iterations.

\section{Related Work}
\textbf{Feature attribution.}  Explainable artificial intelligence (XAI) methods aim to provide human-readable explanations to help users comprehend and trust the outputs created by ML models~\cite{zhang2018visual}. At present, a series of XAI methods have been developed, mainly including building interpretable models (e.g., linear regression model)~\cite{chen2023motif} and generating post hoc explanations. Feature attribution methods (FAMs)~\cite{nielsen2022robust,yang2023idgi,deng2024unifying},  which provide an estimate of the contribution of input features to a model's prediction, is one of the most popular post hoc techniques. In this paper, we restrict our attention to feature attribution.
Existing FAMs mainly measure the contribution of features to prediction through a \emph{perturbation} test, i.e., by perturbing the features and comparing the resulting differences in predictions. 
Typically, Shapley value-based methods~\cite{shapley1953value,lundberg2017unified,zhang2021interpreting,ren2021towards} measure the difference in a model's predictions given data points with and without the feature included. Gradient-based methods~\cite{sundararajan2017axiomatic,yang2023idgi} measure the rate of change in prediction corresponding to a tiny change in an input feature. LIME-based methods~\cite{ribeiro2016should, dhurandhar2022right} learn simplified, interpretable models on perturbed feature subsets. 
In another line, subset-wise FAMs (e.g., GNNExplainer~\cite{ying2019gnnexplainer})  attempt to  find a feature subset with the highest contribution by optimizing a loss function consisting of the $L_p$ norm of a feature mask and the prediction given the masked inputs~\cite{fong2017interpretable,wang2021towards,tan2022learning,lucic2022cf,zhang2023mixupexplainer}. 
However, current FAMs may not accurately distinguish the contributions between different features when their change in prediction is the same after perturbation. To address this issue, we extend feature attribution to compare the differences of each feature in PNS.


\textbf{Probability of Necessity and Sufficiency (PNS). }
Necessity and sufficiency are two different aspects of causation. Sufficiency assesses the probability of a cause actively generating an effect, whereas necessity evaluates whether the cause, when absent, is capable of altering the effect, thereby measuring the effect's dependence on the cause. At present, the main methods for feature attribution that consider sufficiency or necessity are LEWIS~\cite{galhotra2021explaining}, LENS~\cite{watson2021local}, and NSEG~\cite{cai2022probability}. However, since LEWIS and LENS only use PS or PN, the contribution score of an important feature subset may remain unchanged after including redundant features or removing some features inside. For example, we know that `fuel', `oxygen' and `heat' cause `burning'. Since these three key factors and the `ice cream sold out' (redundant feature) can also cause burning, we have PS(fuel, oxygen, heat) = PS(fuel, oxygen, heat, ice cream sold out). Another example is that without `oxygen', `burning' doesn't happen. Thus, PN(fuel, oxygen, heat) = PN(oxygen). NSEG extracts the feature subset with the largest PNS by introducing an identifiability assumption and then optimizing the lower bound of PNS, but our FANS does not require this assumption since we directly compute PNS according to its definition, hinging on the outcome of $\mathbf{Y}$ after the intervention can be observed in our causal model.

\section{Conclusion}
This paper presents a novel attribution method called Feature Attribution with Necessity and Sufficiency (FANS) that can better distinguish the contribution of different feature subsets to predictions. We first formally define the Structural Causality Model (SCM) for the perturbation test, which is widely used by most existing feature attribution methods. 
Building upon this SCM, FANS introduces Necessary and Sufficient Attribution, defined as the highest probability in the Probabilities of being a Necessity and Sufficiency (PNS) cause of the prediction change for  perturbing samples in different neighborhoods,
through our proposed heuristic strategy for estimating the neighborhood and dual-stage (factual and interventional) perturbation test. 
To generate counterfactual samples, we propose to adopt a sampling-importance-resampling approach. Moreover, we show the superior performance of our FANS compared to state-of-the-art methods. The main limitation of our method is that 
the estimation methods of the boundary $b$ of the target input's neighborhood and  threshold $c$ for prediction change are both heuristic.
In future work, it would be meaningful to provide a theoretical framework on the identifiability of $b$ and $c$. 
\clearpage

\section*{Impact Statement}
This paper proposes a feature attribution method called FANS, which can distinguish the contributions of different features even if the results change similarly after perturbation. FANS can be used in medical diagnosis, autonomous driving, and other fields to assist people in understanding the predictions of the model, which have the potential to facilitate the application of existing artificial intelligence technologies in real-world scenarios.

\section*{Acknowledgements}
This research was supported in part by National Key R\&D Program of China (2021ZD0111501), National Science Fund for Excellent Young Scholars (62122022), and Natural Science Foundation of China (62206064, 62206061), and was supported by Turing AI Fellowship under grant EP/V023756/1. We thank Wenlin Chen for valuable suggestions, and the anonymous reviewers who gave useful comments.

\bibliography{citation}
\bibliographystyle{icml2024}

\newpage
\appendix
\onecolumn

\section{Evaluation of PN and PS}\label{sec:derivation}
As illustrated in Figure \ref{fig:process}, the causal diagram demonstrates that variable $\mathbf Y$ is influenced by variables $\tilde{\mathbf X}$ and $\mathbf S$, with a function of $\mathbf Y = f(g(\tilde{\mathbf X}, \mathbf S, \mathbf x'))$ (Eq.~\ref{equ:xs2y}). When variables $\tilde{\mathbf X}$ and $\mathbf S$ are intervened as $\mathbf x$ and $\bar{\mathbf s}$ respectively, according to the counterfactual reasoning paradigm~\cite{pearl2022probabilities}, we have:

$P(\bar{B}_{\mathbf{s}, c_{\bar{A}_{\mathbf{s},b}}} \mid A_{\mathbf{s},b}, B_{\mathbf{s},c}) =\mathbb{E}_{\mathbf{x} \sim P(\tilde{\mathbf{X}} \mid A_{\mathbf{s},b}, B_{\mathbf{s},c})}\left[P\left(\left|\mathbf Y-f(\mathbf{x})\right| \leq c\right)_{\operatorname{do}(\tilde{\mathbf{X}}=\mathbf{x}, \mathbf S = \bar{\mathbf s})}\right]$

For simplicity, we replace $\mathbf Y$ with its value $f(g(\tilde{\mathbf{X}} =\mathbf{x}, \mathbf{S} = \bar{\mathbf{s}}, \mathbf{x}^{\prime}))$, and since $\mathbf S$ is an exogenous variable, the above equation can be written as

$=\mathbb{E}_{\mathbf{x} \sim P(\tilde{\mathbf{X}} \mid A_{\mathbf{s},b}, B_{\mathbf{s},c})}\left[P\left(\left|f\left(g\left(\mathbf{x}, \bar{\mathbf{s}}, \mathbf{x}^{\prime}\right)\right)-f(\mathbf{x})\right| \leq c\right)_{\operatorname{do}(\tilde{\mathbf{X}}=\mathbf{x})}\right]$

Similarly, the derivation of Eq. \ref{equ:ps_def} is given by
\begin{equation*}
\begin{aligned}
    P\left(B_{\mathbf{s},c_{A_{\mathbf{s},b}}} \mid \bar{A}_{\mathbf{s},b}, \bar{B}_{\mathbf{s},c}\right)
    &=\mathbb{E}_{\mathbf{x} \sim P\left(\tilde{\mathbf{X}} \mid \bar{A}_{\mathbf{s},b}, \bar{B}_{\mathbf{s},c}\right)}[P\left(\left|\mathbf Y-f(\mathbf{x})\right|>c\right)_{\mathrm{do}(\tilde{\mathbf{X}}=\mathbf{x}, \mathbf S=\mathbf{s})}]\\
    &=\mathbb{E}_{\mathbf{x} \sim P(\tilde{\mathbf{X}} \mid \bar{A}_{\mathbf{s},b}, \bar{B}_{\mathbf{s},c})}[P\left(\left|f\left(g\left(\mathbf{x}, \mathbf{s}, \mathbf{x}^{\prime}\right)\right)-f(\mathbf{x})\right|>c\right)_{\mathrm{do}(\tilde{\mathbf{X}}=\mathbf{x})}]
\end{aligned}
\end{equation*}

\section{Resampling Weight for Approximate Sampling}\label{sec:sir}
SIR consists of three steps. 
1) Draw samples $\mathbf{x}_1$, ..., $\mathbf{x}_k$ from the proposal distribution $Q(\mathbf{X})$. 
2) For each $\mathbf{x}_i$, 
calculate the weight $w(\mathbf{x}_i) = \frac{P(\mathbf{x}_i|\bar{A}_{\mathbf{s},b},\bar{B}_{\mathbf{s},c})}{Q(\mathbf{x}_i)}$ if $\|\mathbf{x}_{\bar{\mathbf{s}}}-\mathbf{x}^t_{\bar{\mathbf{s}}}\|_p\leq b$ (definition of variable $\tilde{\mathbf{X}}$ in Eq.~\ref{equ:condition}) and $0$ otherwise. 3) Draw samples from $\mathbf{x}_1$, ..., $\mathbf{x}_k$ based on their weights. 

Next, we propose a method for estimating $\frac{P(\mathbf{x}_i|\bar{A}_{\mathbf{s},b},\bar{B}_{\mathbf{s},c})}{Q(\mathbf{x}_i)}$.
As the training set of the ML model comes from $P(\mathbf{X})$, we set $Q(\mathbf{X})$:=$P(\mathbf{X})$, thus 
$w(\mathbf{x}_i)$
=$\frac{P(\mathbf{x}_i| \bar{A}_{\mathbf{s},b},\bar{B}_{\mathbf{s},c})}{P(\mathbf{x}_i)}$
=$\frac{P(\mathbf{x}_i) P(\bar{A}_{\mathbf{s},b},\bar{B}_{\mathbf{s},c}|\mathbf{x}_i)}{P(\mathbf{x}_i)P(\bar{A}_{\mathbf{s},b},\bar{B}_{\mathbf{s},c})}$
=$\frac{ P(\bar{A}_{\mathbf{s},b},\bar{B}_{\mathbf{s},c}|\mathbf{x}_i)}{P(\bar{A}_{\mathbf{s},b},\bar{B}_{\mathbf{s},c})}$
=$r \cdot P(\bar{A}_{\mathbf{s},b},\bar{B}_{\mathbf{s},c}|\mathbf{x}_i)$, 
where $r$ is a normalization constant. Because $\bar A_{\mathbf{s},b}$ denotes an event of perturbation on any dimension subset (except $\mathbf{s}$) of an input $\mathbf{x}$. In our paper, we approximate this event by the complementary feature set $\bar{\mathbf{s}}$, which is represented by the perturbation $g(\mathbf{x}, \bar{\mathbf{s}}, \mathbf{x}')$. $\bar B_{\mathbf{s},c}$ denote an event of the original prediction $f(\mathbf{x})$ remained unchanged relative to $\mathbf{y}$, which is represented by $|\mathbf{y}$$-$$f(\mathbf{x})|$$\le$$c$, where constant $c$ is a hyperparameter. Thus, $P(\bar{A}_{\mathbf{s},b},\bar{B}_{\mathbf{s},c}|\mathbf{x}_i) = P( |f(g(\mathbf{x}_i, \bar{s}, \mathbf{x}')) - f(\mathbf{x})| \le c|\mathbf{x}_i)$.

\begin{table*}[htbp]
  \centering
  \caption{Notation and Descriptions.}
    \begin{tabular}{l|l}
    \toprule
    Notations & Descriptions \\
    \midrule
    $f$ and $g$ & The black box model and perturbation function.\\
    $\mathbf{X}$  & The variable of the input. \\
    $\mathbf{x}$ &  A particular value of $\mathbf{X}$.\\
    $\mathbf{x}'$ & A baseline for perturbation.\\
    $\mathbf{x}^t$,     & Target input to be explained \\
    $\tilde{\mathbf{X}}$     &  Input with fixed features on $\mathbf{S}$ that are similar to $\mathbf{x}^t$. \\
    $\mathbf{Y}$  & The variable of the model's prediction after perturbation. \\
    $\mathbf{y}$ & A particular value of $\mathbf{Y}$.\\ 
    $\mathbf{m}$ & A random mask used to perturb the input.\\
    $\mathbf{S}$   & A set containing the dimensions of $\mathbf X$ specified to be perturbed. $\mathbf{S}$ can be any set and is not fixed.\\
    $\mathbf{s}$  & A particular example of $\mathbf{S}$, indicating the dimensions of the actual perturbation.\\
    $\mathbf{x}_{\mathbf{s}}$, $\mathbf{m}_{\mathbf{s}}$ &  Subvectors of  $\mathbf{x}$ and $\mathbf{m}$ that are indexed by $\mathbf{s}$.\\
    $b$    & Boundary of the neighborhood of the target input (Eq.~\ref{equ:wi}).\\  
    $c$    & Threshold used to determine if there is a significant difference between the perturbed and \\&original predictions.\\ 
    $A_{\mathbf{s},b}$, $\bar A_{\mathbf{s},b}$ &  Perturbation event and its complement.\\
    $B_{\mathbf{s},c}$, $\bar B_{\mathbf{s},c}$ &  Prediction event and its complement.\\
    $r$, $d$  & Normalization constant and the dimensionality of the input. \\ 
    $t$, $u$ & Number of perturbations in the sufficiency and necessity modules, respectively.\\
    $w_{\text{SF}}$ and $w_{\text{NC}}$ & Sample weighting functions for PS and PN, respectively.\\ 
    $\mathcal{E}_{\text{SF}}$ and $\mathcal{E}_{\text{NC}}$ & Sample sets drawn from conditional distributions of PS and PN through SIR, respectively.\\
    $\mathbb I(\cdot)$ and $\circ$ & Indicator function and element-wise multiplication.\\
    \bottomrule
    \end{tabular}%
  \label{tab:notation}%
\end{table*}%

\section{Experimental Setup}
\subsection{Datasets and Target Models}\label{sec:dataset}
\begin{table*}[t!]
  \centering
  \caption{Statistics of the datasets.}
    \begin{tabular}{c|lcccc}
    \toprule
    \multicolumn{1}{c}{Category} & Dataset & \# Images &  Size &    & \# Classes \\
    \midrule
    \multirow{3}[2]{*}{Image} & MNIST & 70,000 & 28$\times$28 &     & 10 \\
          & Fashion-MNIST & 70,000 & 28$\times$28 &     & 10 \\
          & CIFAR10 & 60,000 & 32$\times$32 &     & 10 \\
    \midrule
    \multicolumn{1}{c}{ } &   & \# Nodes & \# Edges & \# Features &   \\
    \midrule
    \multirow{3}[2]{*}{Graph} & BACommunity & 1,400  & 8,598  & 10    & 8 \\
          & Citeseer & 3,327  & 4,732  & 3,703  & 6 \\
          & Pubmed & 19,717 & 44,338 & 500   & 3 \\
    \bottomrule
    \end{tabular}%
  \label{tab:dataset}%
\end{table*}%
To evaluate the effectiveness of our proposed FANS, we utilize six datasets with well-trained black box models to be explained on image and graph domains: (1) on three image datasets (MNIST, Fashion-MNIST, CIFAR10), we train three image classification models using LeNet5 architecture\cite{lecun1998gradient} for the first two datasets and ResNet18\cite{he2016deep} for the third dataset, and (2) on two citation graphs (Citeseer, Pubmed) and one synthetic graph (BACommunity), we train three node classification models, each consisting of three GCN\cite{kipf2016semi} layers, following \cite{wang2021towards}. Table \ref{tab:dataset} summarizes the statistics of six datasets.
\begin{itemize}
    \item \textbf{MNIST}~\cite{lecun1998gradient} contains grayscale images of the 10 digits. 
    \item \textbf{Fashion-MNIST}~\cite{xiao2017fashion} contains grayscale images of the 10 categories. 
    \item \textbf{CIFAR10}~\cite{krizhevsky2009learning} contains color images of the 10 categories. 
    \item \textbf{BACommunity}~\cite{ying2019gnnexplainer} consists of two BA-Shapes graphs and includes eight node categories. A BA-Shapes graph contains a Barabási-Albert (BA) graph and ``house''-structured network motifs.
    \item \textbf{Citeseer} and \textbf{Pubmed}~\cite{sen2008collective} contain documents represented by nodes and citation links represented by edges.
\end{itemize}

\subsection{Baselines}\label{sec:baselines}
We consider two categories of methods, feature-wise attributions and feature subset-wise attributions. Feature-wise attribution methods include:
\begin{itemize}
    \item \textbf{Saliency}~\cite{simonyan2013deep} construct contributions using absolute values of partial derivatives. 
    \item \textbf{GuidedBP}~\cite{DBLP:journals/corr/SpringenbergDBR14} sets the gradients and ReLU inputs to zero if they are negative.
    \item \textbf{IntegratGrad}~\cite{sundararajan2017axiomatic} average gradients along a linear path between the reference and input.
    \item \textbf{GradShap}~\cite{lundberg2017unified} approximates SHAP values by stochastic sampling from the reference distribution and computing the expectations of gradients.
    \item \textbf{DeepLift}~\cite{DBLP:conf/iclr/AnconaCO018}  addresses saturation issues by employing ``reference activations'' computed during the forward pass with the reference input.
    \item \textbf{IDGI}~\cite{yang2023idgi} enhances attribution quality by removing noise in IG-based methods through the focused use of gradients along the important direction.
\end{itemize}
Feature subset-wise attribution methods include:
\begin{itemize}
    \item \textbf{Occlusion}~\cite{zeiler2014visualizing} induces changes in the classifier output through perturbation via sliding a gray square over the input image.
    \item \textbf{LIME}~\cite{ribeiro2016should} quantifies the contributions by using a simplified, interpretable surrogate by fitting the input's neighborhood. 
    \item \textbf{MeaningfulPerturbation(MP)}~\cite{fong2017interpretable}  optimizes the shape of perturbation masks to minimize input image blurring while maximizing class score decrease.
    \item \textbf{GNNExplainer}~\cite{ying2019gnnexplainer}  minimizes the loss by balancing the density penalty and cross-entropy of model prediction on the masked subgraph.
    \item \textbf{PGExplainer}~\cite{luo2020parameterized} extends GNNExplainer by assuming a random Gilbert graph, where edge probabilities are conditionally independent.
    \item \textbf{FeatAblation}~\cite{kokhlikyan2020captum} computes attribution by replacing each input feature with a reference and measuring the output difference.
    \item \textbf{ReFine}~\cite{wang2021towards} involves edge attribution through pre-training, which optimizes mutual information (MI) and contrastive loss, and edge selection through fine-tuning, which solely maximizes MI.
    \item \textbf{CFGNExplainer}~\cite{lucic2022cf} generates counterfactual explanations by learning a perturbed adjacency matrix that flips the classifier prediction for a node.
    \item \textbf{CF$^2$}~\cite{tan2022learning} proposes an objective function for optimizing the edge mask, which consists of factual loss, counterfactual loss, and $L_1$ regularization.
    \item \textbf{NSEG}~\cite{cai2022probability} uses SCMs for counterfactual probability estimation and optimizes continuous masks via gradient updates, using the PNS lower bound as the objective to generate necessary and sufficient GNN explanations.
    \item \textbf{CIMI}~\cite{wu2023causality} minimizes the loss by balancing their proposed sufficiency loss, intervention loss, and causal prior loss.
    \item \textbf{MixUpExplainer}~\cite{zhang2023mixupexplainer} mixes explanations with a randomly sampled base structure to address distribution shifting issues.
\end{itemize}
\subsection{Evaluation Metrics}\label{sec:metric}
It is challenging to evaluate the quality of explanations quantitatively because the ground truth explanations are usually not available. There are six widely used metrics in the literature.
\subsubsection{Infidelity (INF)}
The Infidelity (INF) metric~\cite{yeh2019fidelity} quantifies the expected mean square error between the dot product of an attribution vector and the difference of input and perturbed input, and the difference of output before and after  perturbations. Mathematically, Infidelity is defined as follows.
 \begin{equation*}
 \small
\mathrm{INFD}(\mathbf{s}, \mathbf{x}^t, f) =\mathbb{E}_{\mathbf{m} \sim U(\mathbf{0}, \mathbf{1})}
    \left[\left((\mathbf{x}^t - \mathbf{x}^t \circ \mathbf{m})^T \mathbf{s}-\left(f(\mathbf{x}^t)-f(\mathbf{x}^t \circ \mathbf{m})\right)\right)^2\right],
\end{equation*}
where $U(\mathbf{0}, 
\mathbf{1})$ denotes the uniform distribution over the hypercube $[0, 1]^D$ in $D$-dimensional space, 
$\mathbf{s} \in \mathbb R^D$ represent the attribution vector determined by a FA method.

\subsubsection{IROF (IR)}
 Iterative Removal Of Features  (IROF or IR)~\cite{DBLP:journals/corr/abs-2003-08747} calculates the Area Over the Curve (AOC) for the class score based on the sorted mean importance of feature segments as they are iteratively removed. Formally, IROF is defined as follows.
 \begin{equation*}
     \text{IROF}(\mathbf{a}_1, ..., \mathbf{a}_N, \mathbf{x}^t_1, ..., \mathbf{x}^t_N, f)=\frac{1}{N}\sum_{n=1}^{N}\text{AOC}\left(\frac{f(t(\mathbf{x}^t_n, \mathbf{a}_n, 1))_i}{f(\mathbf{x}^t_n)_i}, ..., \frac{f(t(\mathbf{x}^t_n, \mathbf{a}_n, L))_i}{f(\mathbf{x}^t_n)_i}\right),
 \end{equation*}
where $i$ is a class index, $L$ is the number of segments, $\mathbf x^t_n$ represents the $n$-th target input to be explained, $\mathbf a_n$ denotes the attribution of $\mathbf x^t_n$ predicted by the  feature attribution method to be evaluated. $t(\mathbf x^t_n, \mathbf{a}_n, l)$ is designed to perturb the input  $\mathbf{x}^t_n$ in a segment-wise manner. IROF calculates the importance of each segment in $\mathbf x^t_n$ by averaging the significance scores of individual features within those segments, as provided by its attribution $\mathbf a_n$. The segments are then sorted based on their importance scores in descending order. Function $t$ selectively perturbs the $l$-th ranked segment, yielding the modified input as its output.

\subsubsection{Fidelity$^+$ (FID+) and Fidelity$^-$ (FID-)}
Fidelity$^+$ metric (FID+) and Fidelity$^-$ metric (FID-)~\cite{DBLP:conf/log/AmaraYZHZSBS022} respectively evaluate the proportions of predictions changing and remaining unchanged when features with high attribution scores are removed and retained.
For the sake of convenience in discussion, we assume that $f$ is a multi-class classification model.
\begin{equation}
\small
    \operatorname{Fidelity}^+(\mathbf{s}_1, ..., \mathbf{s}_N, \mathbf{x}^t_1, ..., \mathbf{x}^t_N, f)=1-\frac1N\sum_{i=1}^N \mathbb{I}\left(\arg\max_i f(\mathbf{x}^t_i) = \arg\max_i f(\mathbf{x}^t_i \circ (\mathbf{1} - \mathbf{s})) \right),
\end{equation}
\begin{equation}
\small
    \operatorname{Fidelity}^-(\mathbf{s}_1, ..., \mathbf{s}_N, \mathbf{x}^t_1, ..., \mathbf{x}^t_N, f)=1-\frac1N\sum_{i=1}^N \mathbb{I}\left(\arg\max_i f(\mathbf{x}^t_i) = \arg\max_i f(\mathbf{x}^t_i \circ \mathbf{s}) \right),
\end{equation}
where $\arg\max_i f(\mathbf{x})$ denotes the category with the maximum predicted probability for a given input vector $\mathbf{x}$, where $f(\mathbf{x})$ represents the model's output, typically a vector of class probabilities.

\subsubsection{Max-Sensitivity (MS)}
Max-Sensitivity (MS)~\cite{yeh2019fidelity} computes the maximum difference between new and original attributions, where the new attributions correspond to perturbed inputs.
Given an input neighborhood  radius $r$ and an attribution method $h$, the max-sensitivity is defined as:
\begin{equation}
    \operatorname{SENS}_{\text {MAX }}(h, \mathbf{x}^t, f, r)=\max _{\|\mathbf{z}-\mathbf{x}^t\| \leqslant r} \| h(f, \mathbf{z})-h(f, \mathbf{x}^t)) \|.
\end{equation}

\subsubsection{Sparseness (SPA)}
The Sparseness (SPA) metric~\cite{DBLP:conf/icml/ChalasaniC00J20}  is utilized to quantify the sparsity of an attribution, leveraging the Gini index on the absolute values of the attribution scores. Specifically, given a feature attribution $\mathbf{s}^{\text{sort}}$ sorted in descending order, the Gini Index is given by
\begin{equation}
    \operatorname{Gini}(\mathbf{s}^{\text{sort}})=1-2 \sum_{d=1}^D \frac{s_{d}}{\|\mathbf{s}\|_1}\left(\frac{D-d+0.5}{D}\right)
\end{equation}

\subsubsection{Recall@N}
Recall@N~\cite{wang2021towards} is computed as $\mathbb{E}_{\mathcal{G}}\left[\left|\mathcal{G}_s \cap \mathcal{G}_S^*\right| /\left|\mathcal{G}_S^*\right|\right]$, where $\mathcal{G}^*_s$ is the ground-truth explanatory subgraph. Note that Recall@N in this paper is only suitable for BACommunity since it is a synthetic dataset with known motifs.
\begin{figure}[htb]
\centering
\includegraphics[width=0.5\columnwidth]{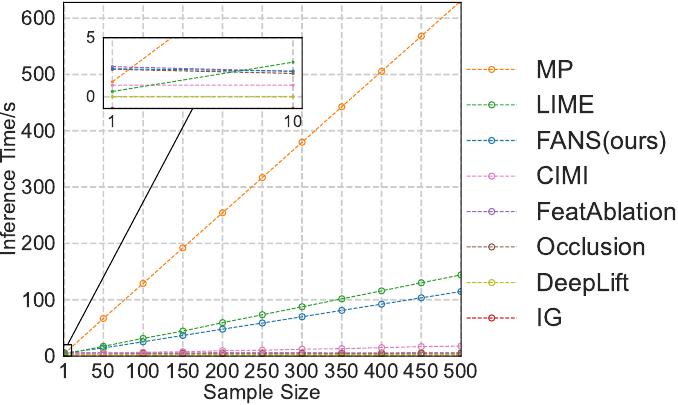}
\vspace*{-1mm}
\caption{Efficiency analysis on the CIFAR10 dataset.}
\label{fig:efficiency}
\vspace*{-3mm}
\end{figure}
\begin{figure}[t!]
	\centering
	\includegraphics[width=0.5\columnwidth]{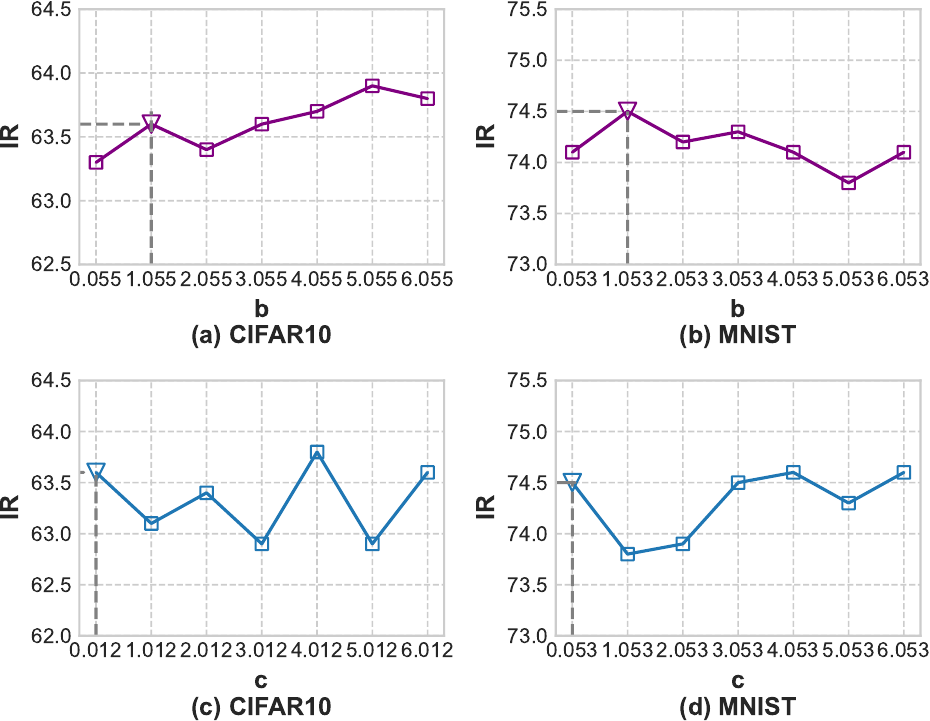} 
 \vspace*{-4mm}
	\caption{Parameter sensitivity analysis for FANS in Eq.~\ref{equ:soft_weight_ps} and Eq.~\ref{equ:soft_weight_pn}.: (a)-(b) boundary $b$ of the neighborhood of the target input $\mathbf{x}^t$. (c)-(d) threshold $c$ for prediction change. The triangle `$\nabla$' represents the values of $b$ or $c$ that we obtained using our heuristic strategy.}
  \vspace*{-5mm}
 \label{fig:parameter}
\end{figure}

\subsection{Implementation Details}\label{sec:imp}
For the  explanation methods with hyperparameters (MP, GNNExplainer, PGExplainer, ReFine, CFGNExplainer, CIMI, MixUpExplainer), we conducted a grid search to tune their hyperparameters. We choose the most commonly used baseline settings based on the type of data. Specifically, for the image dataset, we assign a value to $\mathbf x'$ by sampling from a uniform distribution defined on $[0, 1]^d$. For the graph dataset, we fix $\mathbf{x}'$ to an all-zero vector.

Below are the details of our FANS. Since FANS belong to a feature subset-wise attribution method, we demonstrate the effectiveness of FANS by evaluating the feature subset with the highest PNS estimated by FANS through optimization w.r.t. dimension subset $\mathbf{s}$ (i.e., we adopt the setting of FANS introduced in Section \ref{sec:optimize}). 
In practice, we utilized the Adam optimizer and set the learning rate to 0.001 for image datasets and 0.1 for graph datasets. 
The maximum number of training epochs was set to 50 for image datasets and 30 for graph datasets. 
The times of perturbation $t_{\textsc{SF}}$ in Eq.~\ref{equ:ps_estimate} and $t_{\textsc{NC}}$ in  Eq.~\ref{equ:pn_estimate} were both set to 50.
while the resampling sizes were set to 1. 
According to our proposed strategy in Section \ref{sec:bc}, the values of boundary $b$ on MNIST, Fashion-MNIST, CIFAR10, BACommunity, Citeseer, and Pubmed are set to $1.0539$, $1.0534$, $1.0546$, $1.0599$, $1.0599$ and $1.0600$, respectively.
Correspondingly, the threshold $c$ are set to $0.0534$, $0.0538$, $0.0118$, $7.0710 \times 10^{-6}$, $1.3271 \times 10^{-6}$ and $4.0706 \times 10^{-7}$ for the above datasets, respectively.

\section{Efficiency Analysis}\label{sec:eff}
We evaluate the efficiency of FANS on CIFAR10, in terms of the inference time. We select representative baselines, including optimization-based methods MP, LIME, and CIMI, as well as non-optimization-based methods FeatAblation, Occlusion, DeepLift, and IG. As shown in Figure \ref{fig:efficiency}, the following two main conclusions can be drawn. First, for the same sample size, FANS remains competitive in terms of inference time. Second, for the same time, when the time is approximately greater than 10 seconds, FANS can explain more samples than MP and LIME in the same amount of time. For instance, when given an inference time of 100 seconds, FANS can process approximately 450 samples, whereas LIME and MP can only handle around 350 and 75 samples, respectively. Note that the efficiency of FeatAblation, Occlusion, DeepLift, and IG remains stable, which is expected since they simply compute gradients or perturbations for each sample.

\section{Sensitivity Analysis  of boundary $b$ and threshold $c$}\label{sec:fine_tuning}
We present the sensitivity analysis for the parameters $b$ and $c$ in our FANS. Figure \ref{fig:parameter} (a) and (b) show that the performance of FANS is not sensitive to changes in the values of $b$. In particular, values of $b = 1.055$ on CIFAR10 and $b = 1.053$ on MNIST computed by our proposed heuristic strategy in Section \ref{sec:bc} typically give a robust and comparable performance. For the value of $c$, the performance of FANS gradually decreases and becomes stable, with the value of $c = 0.012$ on CIFAR10 and $c = 0.053$ on MNIST also giving comparable performance.


\end{document}